\documentclass[10pt,twocolumn,letterpaper]{article}

\usepackage{cvpr}
\usepackage{times}
\usepackage{epsfig}
\usepackage{graphicx}
\usepackage{amsmath}
\usepackage{amssymb}
\usepackage{algorithm}
\usepackage{algorithmic}
\usepackage[margin=0pt,font=small,labelfont=small,justification=justified,labelsep=period]{caption}
\usepackage{bm}
\usepackage{multirow}
\usepackage{booktabs}
\usepackage{mathtools}
\usepackage{animate}
\usepackage{xspace}
\usepackage{color}
\usepackage{multirow}
\usepackage{graphics}
\usepackage{adjustbox}
\usepackage{subfigure}
\usepackage{amsmath}
\usepackage{paralist} 
\usepackage{enumitem} 
\usepackage{booktabs} 
\usepackage{array}
\usepackage{grffile}
\usepackage{makecell}

\graphicspath{{figures/}}

\usepackage[pagebackref=true,breaklinks=true,letterpaper=true,colorlinks,bookmarks=false,citecolor=blue,linkcolor=blue]{hyperref}

\newcommand{\Paragraph}[1]{{\vspace{-2mm}\flushleft\textbf{#1}}} 

\long\def\ignorethis#1{}

\usepackage{tabularx}

\definecolor{gray}{rgb}{0.5,0.5,0.5}
\definecolor{MyBlue}{rgb}{0,0,1.0}
\definecolor{MyYellow}{rgb}{0.9,0.9,0}
\definecolor{MyRed}{rgb}{0.8,0.2,0}
\definecolor{MyGreen}{rgb}{0,0.5,0.0}
\definecolor{MyGray}{rgb}{0.4,0.4,0.4}

\def\red#1{\textcolor{MyRed}{#1}}
\def\blue#1{\textcolor{MyBlue}{#1}}
\def\green#1{\textcolor{MyGreen}{#1}}

\def\first#1{\red{\textbf{#1}}}
\def\second#1{\blue{\underline{#1}}}

\newlength\paramargin
\newlength\figmargin
\newlength\secmargin

\newcolumntype{L}[1]{>{\raggedright\let\newline\\\arraybackslash\hspace{0pt}}m{#1}}
\newcolumntype{C}[1]{>{\centering\let\newline\\\arraybackslash\hspace{0pt}}m{#1}}
\newcolumntype{R}[1]{>{\raggedleft\let\newline\\\arraybackslash\hspace{0pt}}m{#1}}

\def\etal{et~al.\xspace}

\newcommand{\figref}[1]{Figure~\ref{fig:#1}}
\newcommand{\tabref}[1]{Table~\ref{tab:#1}}
\newcommand{\eqnref}[1]{\eqref{eq:#1}}

\makeatletter

\setbox0\hbox{$\xdef\scriptratio{\strip@pt\dimexpr
		\numexpr(\sf@size*65536)/\f@size sp}$}

\newcommand{\scriptveryshortarrow}[1][5pt]{{%
		\hbox{\rule[\scriptratio\dimexpr\fontdimen22\textfont2-.2pt\relax]
			{\scriptratio\dimexpr#1\relax}{\scriptratio\dimexpr.4pt\relax}}%
		\mkern-5mu\hbox{\let\f@size\sf@size\usefont{U}{lasy}{m}{n}\symbol{41}}}}

\makeatother

\usepackage[pagebackref=true,breaklinks=true,letterpaper=true,colorlinks,bookmarks=false]{hyperref}

\cvprfinalcopy 

\def\cvprPaperID{1769} 

\def\OurTPAMIs{MEMC-Net}
\def\Ours{DAIN}

\renewcommand{\thefootnote}{\fnsymbol{footnote}}

\begin{document}
	
	\title{Depth-Aware Video Frame Interpolation\vspace{-3mm}}
	
	\author{
		{Wenbo Bao}$^1$
		\hspace{3pt}
		Wei-Sheng Lai$^3$
		\hspace{3pt}
		Chao Ma$^2$
		\hspace{3pt} 
		Xiaoyun Zhang$^{1\ast}$
		\hspace{3pt}
		Zhiyong Gao$^{1}$
		\hspace{3pt}
		Ming-Hsuan Yang$^{3,4}$
		\\
		$^1$ Institute of Image Communication and Network Engineering, Shanghai Jiao Tong University \\
		$^2$ MoE Key Lab of Artificial Intelligence, AI
		Institute, Shanghai Jiao Tong University \\
		$^3$ University of California, Merced
		\hspace{15pt}
		$^4$ Google
		\vspace{-6mm}
	}
	

	\twocolumn[{%
	\renewcommand\twocolumn[1][]{#1}%
	\maketitle
	
\begin{center}
	\begin{minipage}{1\linewidth}
			\footnotesize
			\centering			
			\renewcommand{\tabcolsep}{0.5pt} 
			\renewcommand{\arraystretch}{0.5} 

			\begin{tabular}{lrlrlrlrlr}
				 
				\multicolumn{2}{c}{\includegraphics[width=0.196\linewidth]{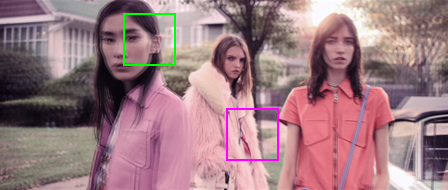}}&
				\multicolumn{2}{c}{\includegraphics[width=0.196\linewidth]{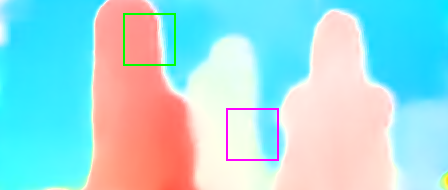}} &
				\multicolumn{2}{c}{\includegraphics[width=0.196\linewidth]{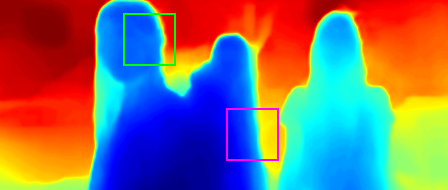}}& 
				\multicolumn{2}{c}{\includegraphics[width=0.196\linewidth]{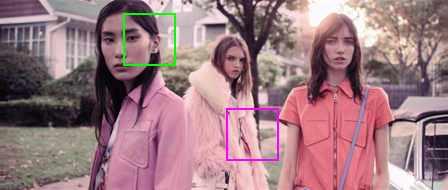}} &
				\multicolumn{2}{c}{\includegraphics[width=0.196\linewidth]{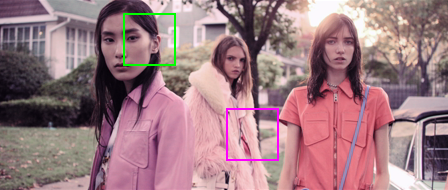}}
				\\
				
				\includegraphics[width=0.098\linewidth]{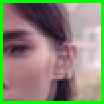} &
				\includegraphics[width=0.098\linewidth]{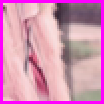} &
				
				\includegraphics[width=0.098\linewidth]{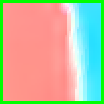} &
				\includegraphics[width=0.098\linewidth]{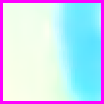} &
				
				\includegraphics[width=0.098\linewidth]{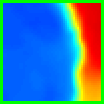} &
				\includegraphics[width=0.098\linewidth]{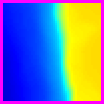} &
				
				\includegraphics[width=0.098\linewidth]{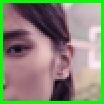} &
				\includegraphics[width=0.098\linewidth]{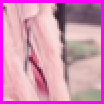} &
				
				\includegraphics[width=0.098\linewidth]{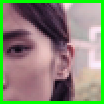} &
				\includegraphics[width=0.098\linewidth]{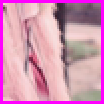} \vspace{1pt} \\
				
	\multicolumn{2}{c}{Overlayed inputs} & \multicolumn{2}{c}{Estimated optical flow} & \multicolumn{2}{c}{Estimated depth map} & \multicolumn{2}{c}{Interpolated frame} & \multicolumn{2}{c}{Ground-truth frame} \\		
				
			\end{tabular} 
\end{minipage}
			\vspace{-5pt}
			\captionof{figure}{
				\textbf{Example of video frame interpolation.}
				We propose a depth-aware video frame interpolation approach to exploit the depth cue for detecting occlusion.
				Our method estimates optical flow with clear motion boundaries and thus generates high-quality frames.
			}
			\label{fig:preface} 
\end{center}
}]

	\thispagestyle{empty}

	\begin{abstract}
		Video frame interpolation aims to synthesize non-existent frames in-between the original frames.
		While significant advances have been made from the recent deep convolutional neural networks, the quality of interpolation is often reduced due to large object motion or occlusion.
		In this work, we propose a video frame interpolation method which explicitly detects the occlusion by exploring the depth information.
		Specifically, we develop a depth-aware flow projection layer to synthesize intermediate flows that preferably sample closer objects than farther ones.
		In addition, we learn hierarchical features to gather contextual information from neighboring pixels.
		The proposed model then warps the input frames, depth maps, and contextual features based on the optical flow and local interpolation kernels for synthesizing the output frame.
		Our model is compact, efficient, and fully differentiable.
		%
		%
		Quantitative and qualitative results demonstrate that the proposed model performs favorably against state-of-the-art frame interpolation methods on a wide variety of datasets.
		The source code and pre-trained model are available at \url{https://github.com/baowenbo/DAIN}.{\let\thefootnote\relax\footnote{$^\ast$Corresponding author}}

	\end{abstract}
	
	\section{Introduction}
	Video frame interpolation has attracted considerable attention in the computer vision community as it can be applied to numerous applications such as slow motion generation~\cite{jiang2017super}, novel view synthesis~\cite{flynn2016deepstereo}, frame rate up-conversion~\cite{bao2018high, castagno1996method}, and frame recovery in video streaming~\cite{wu2016modeling}.
	The videos with a high frame rate can avoid common artifacts, such as temporal jittering and motion blurriness, and therefore are visually more appealing to the viewers.
	However, with the advances of recent deep convolutional neural networks (CNNs) on video frame interpolation~\cite{jiang2017super, liu2017video, niklaus2018context, niklaus2017videoSepConv, xue2017video}, it is still challenging to generate high-quality frames due to large motion and occlusions.

	To handle large motion, several approaches use a coarse-to-fine strategy~\cite{liu2017video} or adopt advanced flow estimation architecture~\cite{niklaus2018context}, e.g., PWC-Net~\cite{sun2018pwc}, to estimate more accurate optical flow.
	On the other hand, a straightforward approach to handle occlusion is to estimate an occlusion mask for adaptively blending the pixels~\cite{bao2018MEMC-Net, jiang2017super, xue2017video}.
	Some recent methods~\cite{niklaus2017videoAdaConv,niklaus2017videoSepConv} learn spatially-varying interpolation kernels to adaptively synthesize pixels from a large neighborhood.
	Recently, the contextual features from a pre-trained classification network have been shown effective for frame synthesis~\cite{niklaus2018context} as the contextual features are extracted from a large receptive field.
	However, all the existing methods rely on a large amount of training data and the model capacity to \emph{implicitly} infer the occlusion, which may not be effective to handle a wide variety of scenes in the wild.

	In this work, we propose to \emph{explicitly} detect the occlusion by exploiting the depth information for video frame interpolation.
	The proposed algorithm is based on a simple observation that closer objects should be preferably synthesized in the intermediate frame.
	Specifically, we first estimate the bi-directional optical flow and depth maps from the two input frames.
	To warp the input frames, we adopt a flow projection layer~\cite{bao2018MEMC-Net} to generate intermediate flows.
	As multiple flow vectors may encounter at the same position, we calculate the contribution of each flow vector based on the depth value for aggregation.
	In contrast to a simple average of flows, the proposed depth-aware flow projection layer generates flows with clearer motion boundaries due to the effect of depth.

	%
	
	Based on our depth-aware flow projection layer, we propose a Depth-Aware video frame INterpolation (DAIN) model that effectively exploits the optical flow, local interpolation kernels, depth maps, and contextual features to synthesize high-quality video frames.
	Instead of relying on a pre-trained recognition network, e.g., ResNet~\cite{he2016deep}, we learn \emph{hierarchical} features to extract effective context information from a large neighborhood.
	We use the adaptive warping layer~\cite{bao2018MEMC-Net} to warp the input frames, contextual features, and depth maps based on the estimated flows and local interpolation kernels.
	Finally, we generate the output frame with residual learning.
	As shown in~\figref{preface}, our model is able to generate frames with clear object shapes and sharp edges.
	Furthermore, the proposed method can generate arbitrary in-between frames for creating slow-motion videos.
	Extensive experiments on multiple benchmarks, including the Middlebury~\cite{baker2011database}, UCF101~\cite{soomro2012ucf101}, Vimeo90K~\cite{xue2017video}, and HD~\cite{bao2018MEMC-Net} datasets, demonstrate that the proposed DAIN performs favorably against existing video frame interpolation methods.

	We make the following contributions in this work:
	\begin{compactitem}
		\item We explicitly detect the occlusion within a depth-aware flow projection layer to preferably synthesize closer objects than farther ones.
		\item We propose a depth-aware video frame interpolation method that tightly integrates optical flow, local interpolation kernels, depth maps, and learnable hierarchical features for high-quality frame synthesis.
		\item We demonstrate that the proposed model is more effective, efficient, and compact than the state-of-the-art approaches.
	\end{compactitem}

	\section{Related Work}
	Video frame interpolation is a long-standing topic and has been extensively studied in the literature~\cite{bao2018high, choi2016map, kim2014new, orchard1994overlapped,wang2010frame}.
	In this section, we focus our discussions on recent learning-based algorithms.
	In addition, we discuss the related topic on depth estimation.

	\Paragraph{Video frame interpolation.}
	As a pioneer of CNN-based methods, Long~\etal~\cite{long2016learning} train a generic CNN to directly synthesize the in-between frame.
	Their results, however, suffer from severe blurriness as a generic CNN is not able to capture the multi-modal distribution of natural images and videos.
	Then, Liu~\etal~\cite{liu2017video} propose the deep voxel flow, a 3D optical flow across space and time, to warp input frames based on a trilinear sampling.
	While the frames synthesized from flow suffer less blurriness, the flow estimation is still challenging for scenes with large motion.
	Inaccurate flow may result in severe distortion and visual artifacts.

	Instead of relying on optical flow, the AdaConv~\cite{niklaus2017videoAdaConv} and SepConv~\cite{niklaus2017videoSepConv} methods estimate spatially-adaptive interpolation kernels to synthesize pixels from a large neighborhood.
	However, these kernel-based approaches typically require high memory footprint and entail heavy computational load.
	Recently, Bao~\etal~\cite{bao2018MEMC-Net} integrate the flow-based and kernel-based approaches into an end-to-end network to inherit the benefit from both sides.
	The input frames are first warped by the optical flow and then sampled via the learned interpolation kernels within an adaptive warping layer.

	Existing methods \emph{implicitly} handle the occlusion by estimating occlusion masks~\cite{bao2018MEMC-Net, jiang2017super, xue2017video}, extracting contextual features~\cite{bao2018MEMC-Net, niklaus2018context}, or learning large local interpolation kernels~\cite{niklaus2017videoAdaConv, niklaus2017videoSepConv}.
	In contrast, we \emph{explicitly} detect the occlusion by utilizing the depth information in the flow projection layer.
	Moreover, we incorporate the depth map with the learned hierarchical features as the contextual information to synthesize the output frame.

	\Paragraph{Depth estimation.}
	Depth is one of the key visual information to understand the 3D geometry of a scene and has been exploited in several recognition tasks, e.g., image segmentation~\cite{zhang2010semantic} and object detection~\cite{sun2010depth}.
	Conventional methods~\cite{ha2016high, karsch2014depth, rajagopalan2004depth} require stereo images as input to estimate the disparity.
	Recently, several learning-based approaches~\cite{eigen2015predicting, eigen2014depth, fu2018deep, kuznietsov2017semi, liu2016learning, roy2016monocular, saxena2006learning, wang2015towards} aim to estimate the depth from a single image.
	In this work, we use the model of Chen~\etal~\cite{chen2016single}, which is an hourglass network trained on the MegaDepth dataset~\cite{li2018megadepth}, for predicting the depth maps from the input frames.
	We show that the initialization of depth network is crucial to infer the occlusion.
	We then jointly fine-tune the depth network with other sub-modules for frame interpolation.
	Therefore, our model learns a \emph{relative} depth for warping and interpolation.

	We note that several approaches jointly estimate optical flow and depth by exploiting the cross-task constraints and consistency~\cite{yin2018geonet, zhou2017unsupervised, zou2018df}.
	While the proposed model also jointly estimates optical flow and depth, our flow and depth are optimized for frame interpolation, which may not resemble the real values of the pixel motion and scene depth.

	\section{Depth-Aware Video Frame Interpolation}
	In this section, we first provide an overview of our frame interpolation algorithm.
	We then introduce the proposed depth-aware flow projection layer, which is the key component to handle occlusion for flow aggregation.
	Finally, we describe the design of all the sub-modules and provide the implementation details of the proposed model.

	\subsection{Algorithm Overview}
	Given two input frames $\mathbf{I}_0(\mathbf{x})$ and $\mathbf{I}_1(\mathbf{x})$, where $\mathbf{x} \in [1,H]\times [1,W]$ indicates the 2D spatial coordinate of the image plane, and $H$ and $W$ are the height and width of the image, our goal is to synthesize an intermediate frame $\hat{\mathbf{I}}_t$ at time $t \in [0, 1]$.
	The proposed method requires optical flows to warp the input frames for synthesizing the intermediate frame.
	We first estimate the bi-directional optical flows, denoted by $\mathbf{F}_{0 \rightarrow 1}$ and $\mathbf{F}_{1 \rightarrow 0}$, respectively.
	To synthesize the intermediate frame $\hat{\mathbf{I}}_t$, there are two common strategies.
	First, one could apply the forward warping~\cite{niklaus2018context} to warp $\mathbf{I}_0$ based on $\mathbf{F}_{0 \rightarrow 1}$ and warp $\mathbf{I}_1$ based on $\mathbf{F}_{1 \rightarrow 0}$.
	However, the forward warping may lead to holes on the warped image.
	The second strategy is to approximate the intermediate flows, i.e., $\mathbf{F}_{t \rightarrow 0}$ and $\mathbf{F}_{t \rightarrow 1}$, and then apply the backward warping to sample the input frames.
	To approximate the intermediate flows, one can borrow the flow vectors from the same grid coordinate in $\mathbf{F}_{0 \rightarrow 1}$ and $\mathbf{F}_{1 \rightarrow 0}$~\cite{jiang2017super}, or aggregate the flow vectors that pass through the same position~\cite{bao2018MEMC-Net}.
	In this work, we adopt the flow projection layer in Bao~\etal~\cite{bao2018MEMC-Net} to aggregate the flow vectors while considering the depth order to detect the occlusion.

	After obtaining the intermediate flows, we warp the input frames, contextual features, and depth maps within an adaptive warping layer~\cite{bao2018MEMC-Net} based on the optical flows and interpolation kernels.
	Finally, we adopt a frame synthesis network to generate the interpolated frame.

	\subsection{Depth-Aware Flow Projection}
\begin{figure}[!t]
	\footnotesize
	\centering	
	\begin{minipage}{0.7\linewidth}
		\centering{ \includegraphics[width= 1.0\textwidth]{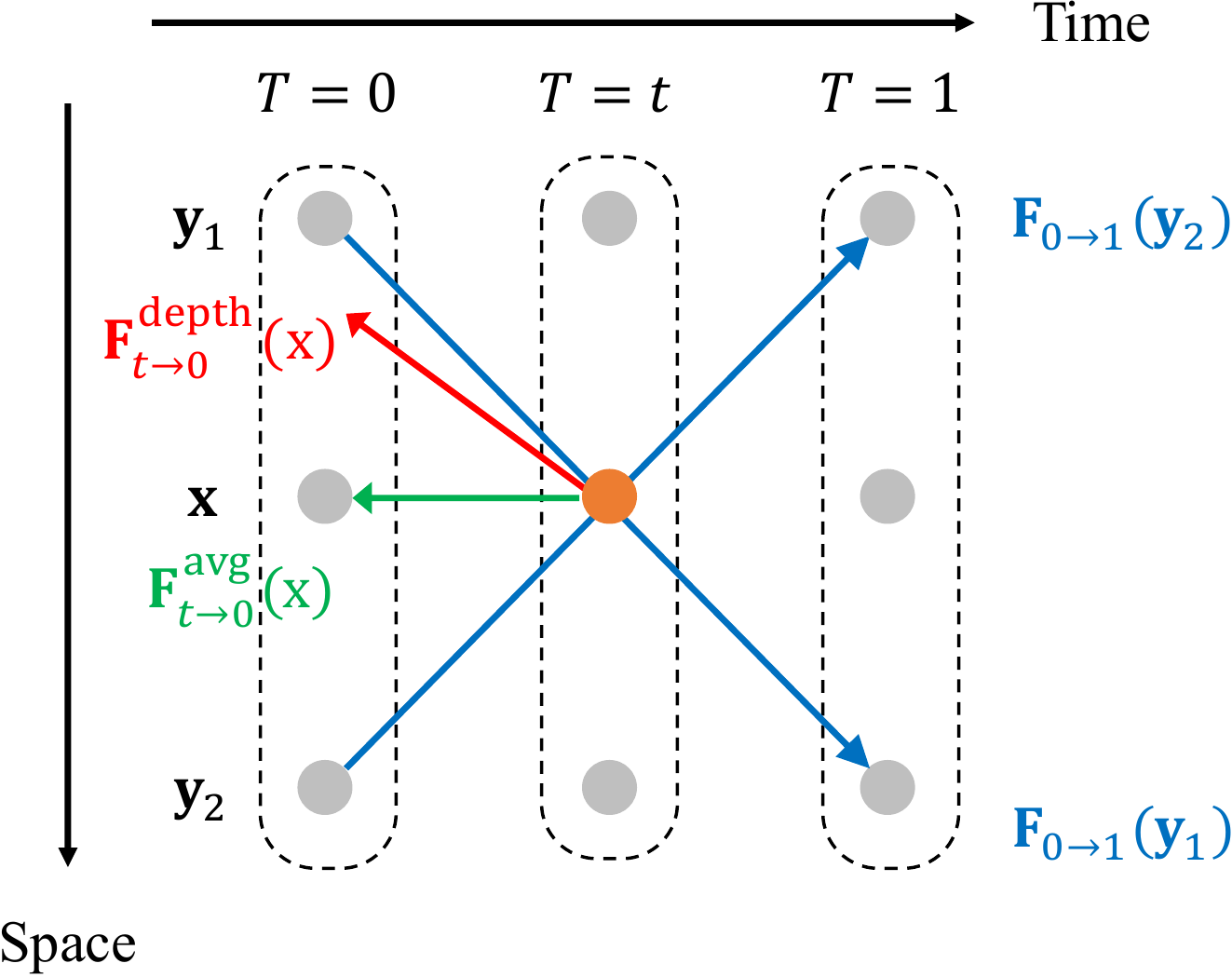}}
	\end{minipage}
	\hfill

	\vspace{-5pt}
	\caption{
		\textbf{Proposed depth-aware flow projection.}
		The existing flow projection method~\cite{bao2018MEMC-Net} obtains an average flow vector which may not point to the correct object or pixel.
		In contrast, we re-write the flows according to the depth values and generate the flow vector pointing to the closer pixel.
		%
		%
	}
 	\vspace{-10pt}
	\label{fig:flow-project} 
\end{figure}

	The flow projection layer approximates the intermediate flow at a given position $\mathbf{x}$ by ``reversing'' the flow vectors passing through $\mathbf{x}$ at time $t$.
	If the flow $\mathbf{F}_{0 \rightarrow 1}({\mathbf{y}})$ passes through $\mathbf{x}$ at time $t$, one can approximate $\mathbf{F}_{t \rightarrow 0}(\mathbf{x})$ by $-t~\mathbf{F}_{0 \rightarrow 1} (\mathbf{y})$.
	Similarly, we approximate $\mathbf{F}_{t \rightarrow 1}(\mathbf{x})$ by $-(1-t)~\mathbf{F}_{1 \rightarrow 0} (\mathbf{y})$.
	However, as illustrated in the 1D space-time example of~\figref{flow-project}, multiple flow vectors could be projected to the same position at time $t$.
	Instead of aggregating the flows by a simple average~\cite{bao2018MEMC-Net}, we propose to consider the depth ordering for aggregation.
	Specifically, we assume that $D_0$ is the depth map of $\mathbf{I}_0$ and $\mathcal{S}(\mathbf{x}) = \big\{\mathbf{y}: \mathrm{round} (\mathbf{y} + t~\mathbf{F}_{0 \rightarrow 1}(\mathbf{y})) = \mathbf{x}, \forall\ \mathbf{y} \in [1,H]\times[1,W]  \big\}$ indicates the set of pixels that pass through the position $\mathbf{x}$ at time $t$.
	The projected flow $\mathbf{F}_{t \rightarrow 0}$ is defined by:
	\begin{equation}
	\mathbf{F}_{t \rightarrow 0}(\mathbf{x}) = 
	-t \cdot
	\frac{ \sum\limits_{\mathbf{y} \in \mathcal{S}(\mathbf{x})} w_0(\mathbf{y}) \cdot \mathbf{F}_{0 \rightarrow 1} (\mathbf{y}) }
	{ \sum\limits_{\mathbf{y} \in \mathcal{S}(\mathbf{x})} { w_0 (\mathbf{y})} },
	\label{eq:flow_project}
	\end{equation}
	where the weight $w_0$ is the reciprocal of depth:
	\begin{equation}
	w_0(\mathbf{y}) = \frac{1}{D_0(\mathbf{y})}.
	\end{equation}
	Similarly, the projected flow $\mathbf{F}_{t \rightarrow 1}$ can be obtained from the flow $\mathbf{F}_{1 \rightarrow 0}$ and depth map $D_1$.
	By this way, the projected flows tend to sample the closer objects and reduce the contribution of occluded pixels which have larger depth values.
	As shown in~\figref{flow-project}, the flow projection used in~\cite{bao2018MEMC-Net} generates an average flow vector (the \green{\textbf{green}} arrow), which may not point to the correct pixel for sampling.
	In contrast, the projected flow from our depth-aware flow projection layer (the \red{\textbf{red}} arrow) points to the pixel with a smaller depth value.
	%

	On the other hand, there might exist positions where none of the flow vectors pass through, leading to holes in the intermediate flow.
	To fill in the holes, we use the outside-in strategy~\cite{baker2011database}: the flow in the hole position is computed by averaging the available flows from its neighbors:
	\begin{equation}
	\mathbf{F}_{t \rightarrow 0}(\mathbf{x}) =
	\frac{1}{ |\mathcal{N}(\mathbf{x})| }
	\sum_{\mathbf{x}^{\prime} \in \mathcal{N}(\mathbf{x})} \mathbf{F}_{t \rightarrow 0}(\mathbf{x}^{\prime}),
	\label{eq:flow_project_hole}
	\end{equation}
	where $\mathcal{N}(\mathbf{x}) = \{\mathbf{x}^{\prime}: |\mathcal{S}(\mathbf{x}^{\prime})| > 0 \}$ is the 4-neighbors of $\mathbf{x}$. 
	From~\eqnref{flow_project} and~\eqnref{flow_project_hole}, we obtain dense intermediate flow fields $\mathbf{F}_{t \rightarrow 0}$ and $\mathbf{F}_{t \rightarrow 1}$ for warping the input frames.

	The proposed depth-aware flow projection layer is fully differentiable so that both the flow and depth estimation networks can be jointly optimized during the training.
	We provide the details of back-propagation in depth-aware flow projection in the supplementary materials.

	\begin{figure*}
	\footnotesize
	\centering
	\includegraphics[width=1.0\linewidth]{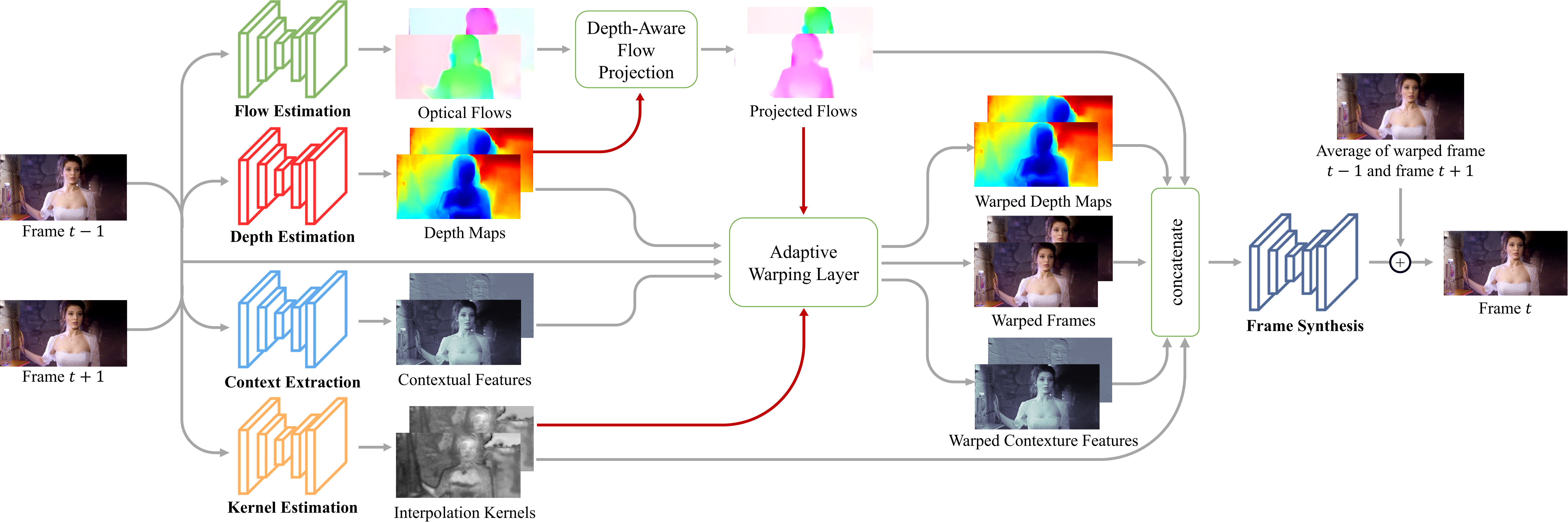}%
	 \vspace{-5pt}
	\caption{
		\textbf{Architecture of the proposed depth-aware video frame interpolation model.}
		Given two input frames, we first estimate the optical flows and depth maps and use the proposed depth-aware flow projection layer to generate intermediate flows.
		We then adopt the adaptive warping layer to warp the input frames, depth maps, and contextual features based on the flows and spatially varying interpolation kernels.
		Finally, we apply a frame synthesis network to generate the output frame.
	}
	\vspace{-10pt}
	\label{fig:netarch}
\end{figure*}

	\subsection{Video Frame Interpolation}
	The proposed model consists of the following sub-modules: the flow estimation, depth estimation, context extraction, kernel estimation, and frame synthesis networks.
	We use the proposed depth-aware flow projection layer to obtain intermediate flows and then warp the input frames, depth maps, and contextual features within the adaptive warping layer.
	Finally, the frame synthesis network generates the output frame with residual learning.
	We show the overall network architecture in~\figref{netarch}.
	Below we describe the details of each sub-network.

	\Paragraph{Flow estimation.}
	We adopt the state-of-the-art flow model, PWC-Net~\cite{sun2018pwc}, as our flow estimation network.
	As learning optical flow without ground-truth supervision is extremely difficult, we initialize our flow estimation network from the pre-trained PWC-Net.

	\Paragraph{Depth estimation.}
	We use the hourglass architecture~\cite{chen2016single} as our depth estimation network.
	To obtain meaningful depth information for the flow projection, we initialize the depth estimation network from the pre-trained model of Li~\etal~\cite{li2018megadepth}.

	\begin{figure}[!t]	 
	\footnotesize
	\centering
	\begin{tabular}{cc}
		\includegraphics[height=0.55\linewidth]{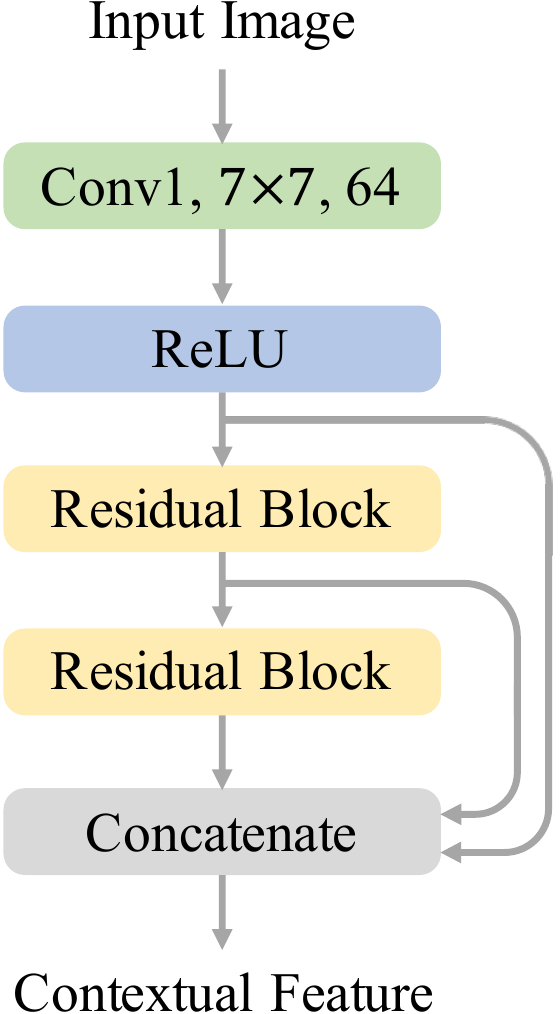}
		&
		\includegraphics[height=0.55\linewidth]{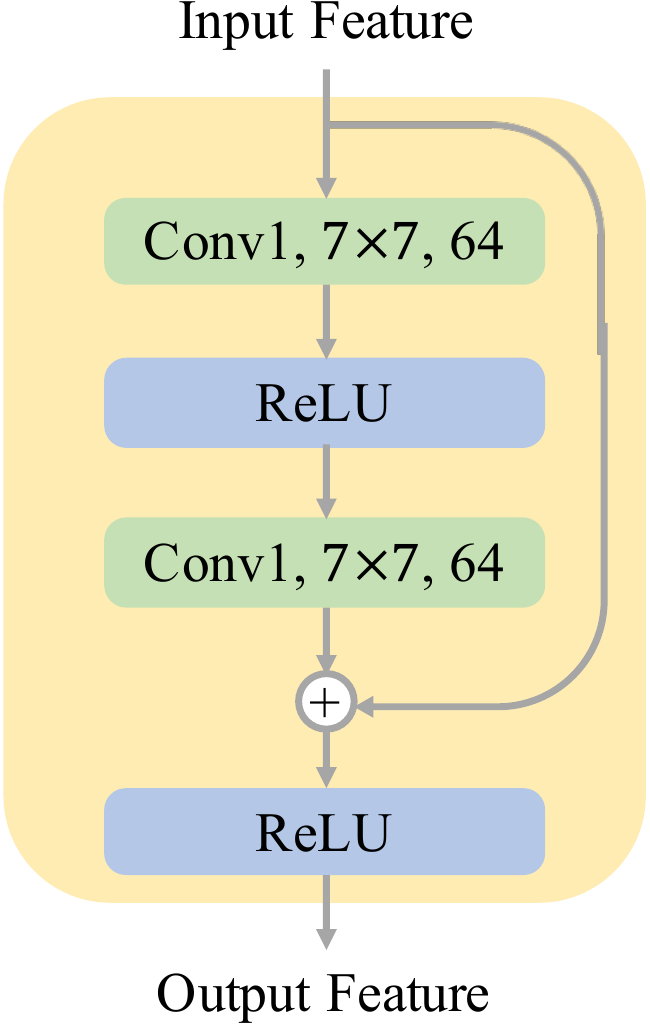}
		\\
		(a) Context extraction network & 
		(b) Residual block \\
	\end{tabular}
	\vspace{-5pt}
	\caption{%
		\textbf{Structure of the context extraction network.}
		Instead of using the weights of a pre-trained classification network~\cite{niklaus2018context}, we train our context extraction network from scratch and learn hierarchical features for video frame interpolation.
	}
	\label{fig:context-extraction}
 	\vspace{-10pt} 
\end{figure}

	\Paragraph{Context extraction.}
	In~\cite{bao2018MEMC-Net} and~\cite{niklaus2018context}, the contextual information is extracted by a pre-trained ResNet~\cite{he2016deep}, i.e., the feature maps of the first convolutional layer.
	However, the features from the ResNet are for the image classification task, which may not be effective for video frame interpolation.
	Therefore, we propose to \emph{learn} the contextual features.
	Specifically, we construct a context extraction network with one $7 \times 7$ convolutional layer and two residual blocks, as shown in~\figref{context-extraction}(a).
	The residual block consists of two $3 \times 3$ convolutional and two ReLU activation layers (\figref{context-extraction}(b)).
	We do not use any normalization layer, e.g., batch normalization.
	We then concatenate the features from the first convolutional layer and the two residual blocks, resulting in a \emph{hierarchical} feature.
	Our context extraction network is trained from scratch and, therefore, learns effective contextual features for video frame interpolation.

	\Paragraph{Kernel estimation and adaptive warping layer.}
	The local interpolation kernels have been shown to be effective for synthesizing a pixel from a large local neighborhood~\cite{niklaus2017videoAdaConv, niklaus2017videoSepConv}.
	Bao~\etal~\cite{bao2018MEMC-Net} further integrate the interpolation kernels and optical flow within an adaptive warping layer.
	The adaptive warping layer synthesizes a new pixel by sampling the input image within a local window, where the center of the window is specified by optical flow.
	Here we use a U-Net architecture~\cite{ronneberger2015u} to estimate $4 \times 4$ local kernels for each pixel.
	With the interpolation kernels and intermediate flows generated from the depth-aware flow projection layer, we adopt the adaptive warping layer~\cite{bao2018MEMC-Net} to warp the input frames, depth maps, and contextual features.
	More details of the adaptive warping layer and the configuration of the kernel estimation network are provided in the supplementary materials.

	\Paragraph{Frame synthesis.}
	To generate the final output frame, we construct a frame synthesis network, which consists of 3 residual blocks.
	We concatenate the warped input frames, warped depth maps, warped contextual features, projected flows, and interpolation kernels as the input to the frame synthesis network.
	In addition, we linearly blend the two warped frames and enforce the network to predict the residuals between the ground-truth frame and the blended frame.
	We note that the warped frames are already aligned by the optical flow.
	Therefore, the frame synthesis network focuses on enhancing the details to make the output frame look sharper.
	We provide the detailed configurations of the frame synthesis network in the supplementary material.

	\subsection{Implementation Details}
	\Paragraph{Loss Function.}
	We denote the synthesized frame by $\hat{\mathbf{I}}_t$ and the ground-truth frame by $\mathbf{I}^{GT}_t$.
	We train the proposed model by optimizing the following loss function:
	\begin{equation}
	\mathcal{L} = \sum_{\mathbf{x}} \rho\left( \hat{\mathbf{I}}_t(\mathbf{x}) - \mathbf{I}^{GT}_{t}(\mathbf{x}) \right), 
	\label{eq:loss}
	\end{equation}
	where $\rho(x) = \sqrt{x^2 +\epsilon^2}$ is the Charbonnier penalty function~\cite{charbonnier1994two}.
	We set the constant $\epsilon$ to $1e-6$.

	\Paragraph{Training Dataset.}
	We use the Vimeo90K dataset~\cite{xue2017video} to train our model.
	The Vimeo90K dataset has 51,312 triplets for training, where each triplet contains 3 consecutive video frames with a resolution of $256 \times 448$ pixels.
	We train our network to predict the middle frame (i.e., $t = 0.5$) of each triplet.
	At the test time, our model is able to generate arbitrary intermediate frames for any $t \in [0, 1]$.
	We augment the training data by horizontal and vertical flipping as well as reversing the temporal order of the triplet.

	\Paragraph{Training Strategy.}
	We use the AdaMax~\cite{kingma2015adam} to optimize the proposed network.
	We set the $\beta_1$ and $\beta_2$ to 0.9 and 0.999 and use a batch size of 2.
	The initial learning rates of the kernel estimation, context extraction, and frame synthesis networks are set to $1e-4$.
	As both the flow estimation and depth estimation networks are initialized from pre-trained models, we use smaller learning rates of $1e-6$ and $1e-7$, respectively.
	We jointly train the entire model for 30 epochs and then reduce the learning rate of each network by a factor of 0.2 and fine-tune the entire model for another 10 epochs.
	We train our model on an NVIDIA Titan X (Pascal) GPU card, which takes about 5 days to converge.

	\section{Experimental Results}
	In this section, we first introduce the datasets for evaluation.
	We then conduct ablation study to analyze the contribution of the proposed depth-aware flow projection and hierarchical contextual features.
	Then, we compare the proposed model with state-of-the-art frame interpolation algorithms.
	Finally, we discuss the limitation and future work of our method.

	\subsection{Evaluation Datasets and Metrics}
	We evaluate the proposed algorithm on multiple video datasets with different image resolutions.
	
	\Paragraph{Middlebury}. The Middlebury benchmark~\cite{baker2011database} is widely used to evaluate video frame interpolation methods.
	There are two subsets.
	The \textsc{Other} set provides the ground-truth middle frames, while the \textsc{Evaluation} set hides the ground-truth and can be evaluated by uploading the results to the benchmark website.
	The image resolution in this dataset is around $640 \times 480$ pixels.

	\Paragraph{Vimeo90K}.
	There are 3,782 triplets in the test set of the Vimeo90K dataset~\cite{xue2017video}.
	The image resolution in this dataset is $448 \times 256$ pixels.

	\Paragraph{UCF101.}
	The UCF101 dataset~\cite{soomro2012ucf101} contains videos with a large variety of human actions. 
	There are 379 triplets with a resolution of $256 \times 256$ pixels.

	\Paragraph{HD.} 
	Bao~\etal~\cite{bao2018MEMC-Net} collect 11 high-resolution videos for evaluation.
	The HD dataset consists of four $1920\times 1080$p, three $1280\times 720$p and four $1280 \times 544$p videos.
	The motion in this dataset is typically larger than other datasets.

	\Paragraph{Metrics.}
	We compute the average Interpolation Error (IE) and Normalized Interpolation Error (NIE) on the Middlebury dataset.
	Lower {IE}s or {NIE}s indicate better performance.
	We evaluate the PSNR and SSIM on the UCF101, Vimeo90K, and the HD datasets for comparisons.

	\begin{table}
	\caption{
		\textbf{Analysis on Depth-Aware (DA) flow projection.} 
		M.B. is short for the \textsc{Other} set of the Middlebury dataset.
		The proposed model (DA-\textit{Opti}) shows a substantial improvement against the other variations.
	}
	\vspace{-1mm}
	\label{tab:depth-aware_ablation}
	\footnotesize
	\renewcommand{\tabcolsep}{3pt} 
	\centering
	\begin{tabular}{cccccccc}
		\toprule
		\multirow{2}{*}[-0.28em]{Method} &  
		\multicolumn{2}{c}{UCF101~\cite{soomro2012ucf101}} &
		\multicolumn{2}{c}{Vimeo90K~\cite{xue2017video}} &
		M.B.~\cite{baker2011database} &
		\multicolumn{2}{c}{HD~\cite{bao2018MEMC-Net}}\\
		\cmidrule(l{2pt}r{1pt}){2-3}
		\cmidrule(l{2pt}r{1pt}){4-5} 
		\cmidrule(l{2pt}r{1pt}){6-6}
		\cmidrule(l{2pt}r{1pt}){7-8}
		
		&PSNR & SSIM 	&PSNR & SSIM	& 
		IE & PSNR & SSIM \\
		\midrule
		DA-\textit{{None}}	 
		&\second{34.91}  &0.9679   &34.47 &0.9746 & 2.10 & 31.46 & 0.9174 \\

		DA-\textit{{Scra}} 
		&34.85  &  0.9677 & 34.30 &0.9735 & 2.13& 31.42 & 0.9164\\

		DA-\textit{{Pret}} 
		& \second{34.91} & \second{0.9680}  & \second{34.52} & \second{0.9747} & \second{2.07} & \second{31.52} & \second{0.9178}\\		
	
	 	DA-\textit{{Opti}} 
		&\first{34.99}  &  \first{0.9683} & \first{34.71} &\first{0.9756} &\first{2.04} &\first{31.70} & \first{0.9193}\\
		\bottomrule
	\end{tabular}
		\vspace{-10pt} 	
\end{table}

	\begin{figure}
	\footnotesize
	\centering
	\renewcommand{\tabcolsep}{1pt} 
	\renewcommand{\arraystretch}{0.7} 
		\begin{tabular}{cccc}
			\parbox[t]{2mm}{\multirow{1}{*}[4em]{\rotatebox[origin=c]{90}{DA-\textit{\footnotesize{None}}}}} &
			&
			\includegraphics[width=0.32\linewidth]{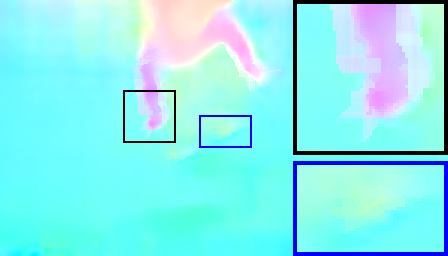}&
			\includegraphics[width=0.32\linewidth]{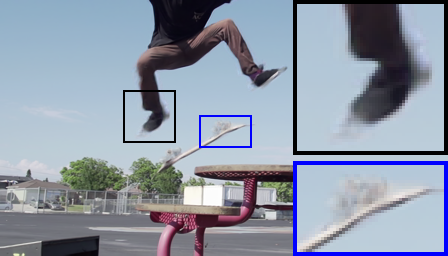}
			\\
			\parbox[t]{2mm}{\multirow{1}{*}[4em]{\rotatebox[origin=c]{90}{DA-\textit{\footnotesize{Scra}}}}} &
			\includegraphics[width=0.32\linewidth]{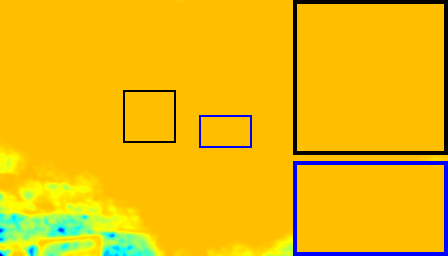}&
			\includegraphics[width=0.32\linewidth]{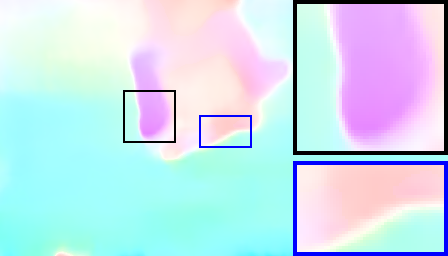}&
			\includegraphics[width=0.32\linewidth]{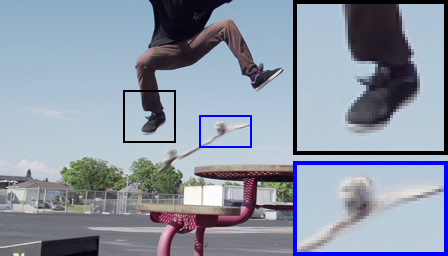}
			\\
			
			\parbox[t]{2mm}{\multirow{1}{*}[4em]{\rotatebox[origin=c]{90}{DA-\textit{\footnotesize{Pret}}}}} &
			\includegraphics[width=0.32\linewidth]{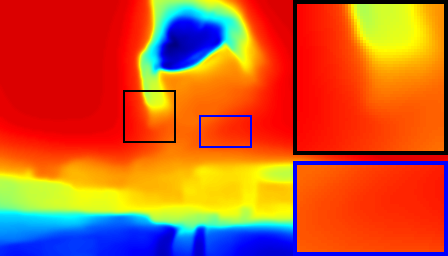}&
			\includegraphics[width=0.32\linewidth]{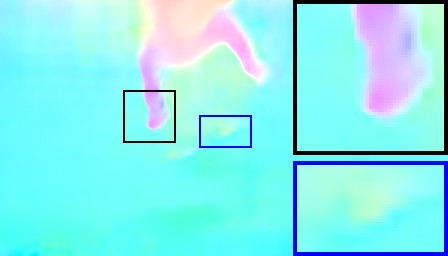}&
			\includegraphics[width=0.32\linewidth]{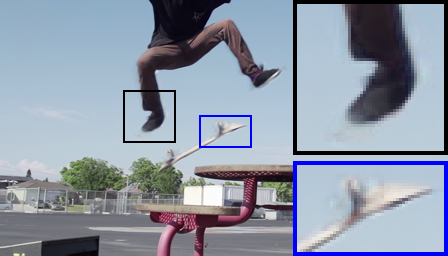}
			\\
			
			\parbox[t]{2mm}{\multirow{1}{*}[4em]{\rotatebox[origin=c]{90}{DA-\textit{\footnotesize{Opti}}}}} &
			\includegraphics[width=0.32\linewidth]{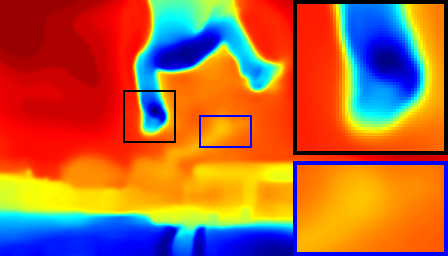}&
			\includegraphics[width=0.32\linewidth]{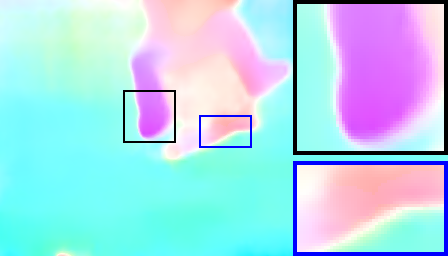}&
			\includegraphics[width=0.32\linewidth]{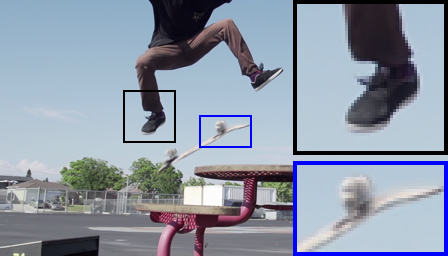}
			\\
			\vspace{1mm}
			&
			Depth map&
			Optical flow&
			Interpolated frame \\
		\end{tabular}
	\vspace{-5pt}
	\caption{
		\textbf{Effect of the depth-aware flow projection.} 
		The DA-\textit{Scra} model cannot learn any meaningful depth information.
		The DA-\textit{Pret} model initializes the depth estimation network from a pre-trained model and generates clear motion boundaries for frame interpolation.
		The DA-\textit{Opti} model further optimizes the depth maps and generates sharper edges and shapes.
	}
	\label{fig:effect_of_depth_aware} 
	\vspace{-10pt}
\end{figure}

	\subsection{Model Analysis}
	\label{sec:Ablation}
	We analyze the contribution of the two key components in the proposed model: the depth-aware flow projection layer and learned hierarchical contextual features.

	\Paragraph{Depth-aware flow projection.}
	To analyze the effectiveness of our depth-aware flow projection layer, we train the following variations (DA is short for Depth-Aware):
	\begin{compactitem}
		\item DA-\textit{{None}}: We remove the depth estimation network and use a simple average~\cite{bao2018MEMC-Net} to aggregate the flows in the flow projection layer.
		\item DA-\textit{{Scra}}: We initialize the depth estimation network from scratch and optimize it with the whole model.
		\item DA-\textit{{Pret}}: We initialize the depth estimation network from the pre-trained model of~\cite{li2018megadepth} but freeze the parameters.
		\item DA-\textit{{Opti}}: We initialize the depth estimation network from the pre-trained model of~\cite{li2018megadepth} and jointly optimize it with the entire model.
	\end{compactitem}

	We show the quantitative results of the above models in~\tabref{depth-aware_ablation} and provide a visualization of the depth, flow, and interpolated frames in~\figref{effect_of_depth_aware}.
	First, the DA-\textit{Scra} model performs worse than the DA-\textit{None} model.
	As shown in the second row of~\figref{effect_of_depth_aware}, the DA-\textit{Scra} model cannot learn any meaningful depth information from the random initialization.
	When initializing from the pre-trained depth model, the DA-\textit{Pret} model shows a substantial performance improvement and generates flow with clear motion boundaries.
	After jointly optimizing the whole network, the DA-\textit{Opti} model further improves the depth maps, e.g., the man's legs, and generates sharper edges for the shoes and skateboard in the interpolated frame.
	The analysis demonstrates that the proposed model effectively utilizes the depth information to generate high-quality results.

	\Paragraph{Learned hierarchical context}.
	In the proposed model, we use contextual features as one of the inputs to the frame synthesis network.
	We analyze the contribution of the different contextual features, including the pre-trained \textit{conv1} features (PCF), the learned \textit{conv1} features (LCF), and the learned hierarchical features (LHF).
	In addition, we also consider the depth maps (D) as the additional contextual features.

	\begin{table}
	\caption{
		\textbf{Analysis on contextual features.}
		We compare the contextual features from different sources: the pre-trained \textit{conv1} features (PCF), learned \textit{conv1} features (LCF), learned hierarchical features (LHF), and the depth maps (D).
	}
	\vspace{-5pt}
	\label{tab:context_ablation}
	\footnotesize
	\renewcommand{\tabcolsep}{3pt} 
	\centering
	\begin{tabular}{cccccccc}
		\toprule
		\multirow{2}{*}[-0.28em]{Context} &
		\multicolumn{2}{c}{UCF101~\cite{soomro2012ucf101}} &
		\multicolumn{2}{c}{Vimeo~\cite{xue2017video}} &
		M.B.~\cite{baker2011database} &
		\multicolumn{2}{c}{HD~\cite{bao2018MEMC-Net}}
		\\
		\cmidrule(l{2pt}r{1pt}){2-3}
		\cmidrule(l{2pt}r{1pt}){4-5} 
		\cmidrule(l{2pt}r{1pt}){6-6}
		\cmidrule(l{2pt}r{1pt}){7-8}
		
		&PSNR & SSIM 	&PSNR & SSIM	& 
		IE &PSNR & SSIM \\
		\addlinespace[-1pt]
		\midrule
		
		None 
		&34.84 & 0.9679 & 34.38 &0.9738 & 2.21 & 31.35 & 0.9178 \\
		
		PCF
		&{34.90} & 0.9681 & 34.41 &0.9740 & 2.16 & 31.43 & 0.9160 \\
		
		D
		&{34.90}  &\second{0.9682} &34.44 &0.9740 &  2.14 &31.62 & 0.9183\\ 
		
		PCF + D
		&\second{34.97} & \second{0.9682} &34.49 & 0.9746 & 2.13 &\first{31.73}  & \first{0.9194}\\	
		
		LCF + D
		& 34.87 & 0.9680 & \second{34.54} &  \second{0.9749} & \second{2.08} & 31.56  & 0.9185 \\	 
		
		LHF + D
		&\first{34.99}  &  \first{0.9683} & \first{34.71} &\first{0.9756} & \first{2.04} & \second{31.70} & \second{0.9193}\\		
		\bottomrule
	\end{tabular} 	
\end{table}

	\begin{figure}
	\scriptsize 
	\centering
	\renewcommand{\tabcolsep}{1pt} 
	\begin{tabular}{ccccccc}
			GT & None & PCF & D & PCF + D & LCF + D & LHF + D
			\\
			\includegraphics[width=0.13\linewidth]{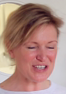} &
			\includegraphics[width=0.13\linewidth]{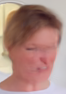}&
			\includegraphics[width=0.13\linewidth]{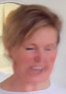} &
			\includegraphics[width=0.13\linewidth]{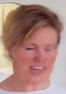}&			
			\includegraphics[width=0.13\linewidth]{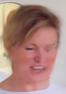}&
			\includegraphics[width=0.13\linewidth]{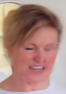}&
			\includegraphics[width=0.13\linewidth]{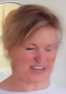}
			\\
			
			PSNR & 29.28 &30.70 & 30.67 & 31.22 & 31.49 &31.55\\
			
			\includegraphics[width=0.13\linewidth]{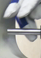} &
			\includegraphics[width=0.13\linewidth]{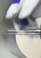}&
			\includegraphics[width=0.13\linewidth]{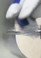}&
			\includegraphics[width=0.13\linewidth]{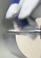}&
			\includegraphics[width=0.13\linewidth]{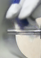} &	
			\includegraphics[width=0.13\linewidth]{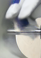} &
			\includegraphics[width=0.13\linewidth]{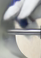}
			\\
			PSNR & 30.86&31.01 &31.11 &31.28   & 31.31 & 31.66 \\
		\end{tabular}
	\vspace{-5pt}
	\caption{
		\textbf{Effect of contextual features.} 		
		The proposed model uses the learned hierarchical features (LHF) and depth maps (D) for frame synthesis, which generates clearer and sharper content.
	}
	\label{fig:effect_of_context} 
	\vspace{-10pt}
\end{figure}

	\begin{table*} 
 \caption{
 	\textbf{Quantitative comparisons on the Middlebury \textsc{Evaluation} set.} 
 	The numbers in \first{red} and \second{blue} represent the best and second best performance.
	The proposed DAIN method performs favorably against other approaches in terms of IE and NIE.
 	}
 \label{tab:Middlebury}
 \vspace{-2mm}
\footnotesize
\renewcommand{\tabcolsep}{5pt} 
\renewcommand{\arraystretch}{1} 
\centering
\begin{tabular}{rcccccccccccccccc|cc}
\toprule
\multirow{2}{*}[-0.28em]{Method}& \multicolumn{2}{c}{Mequon} & \multicolumn{2}{c}{Schefflera} & \multicolumn{2}{c}{Urban} & \multicolumn{2}{c}{Teddy} & \multicolumn{2}{c}{Backyard} & \multicolumn{2}{c}{Basketball} & \multicolumn{2}{c}{Dumptruck} & \multicolumn{2}{c}{Evergreen} & \multicolumn{2}{c}{{Average}}\\
\cmidrule(l{2pt}r{2pt}){2-3}
\cmidrule(l{2pt}r{2pt}){4-5}
\cmidrule(l{2pt}r{2pt}){6-7}
\cmidrule(l{2pt}r{2pt}){8-9}
\cmidrule(l{2pt}r{2pt}){10-11}
\cmidrule(l{2pt}r{2pt}){12-13}
\cmidrule(l{2pt}r{2pt}){14-15}
\cmidrule(l{2pt}r{2pt}){16-17}
\cmidrule(l{2pt}r{2pt}){18-19}
&IE & NIE&IE & NIE&IE & NIE&IE & NIE&IE & NIE&IE & NIE&IE & NIE&IE & NIE&IE & NIE\\
\addlinespace[-1pt]
\midrule

EpicFlow~\cite{revaud2015epicflow} 		    &3.17	&0.62 	&3.79    &0.70      &4.28        &1.06      &6.37     &1.09 	 &11.2      &1.18 &6.23   &1.10    &8.11 &1.00    	&8.76      &1.04  &6.49   &0.97  \\

SepConv-$L_1$~\cite{niklaus2017videoSepConv} &{2.52} 	 &\second{0.54}  &3.56 	 &0.67   	&4.17 		 &1.07  	&5.41 	  &	1.03 	&10.2 & 	0.99 &5.47 	 &	{0.96}  	&6.88 & 0.68  	&{6.63} & 	0.70 	&5.61   &	0.83 	 \\

ToFlow~\cite{xue2017video} &2.54 &0.55 & 3.70 & 0.72 & 3.43 &0.92 &5.05 &0.96 & 9.84 &0.97 & 5.34 & 0.98 & 6.88 & 0.72 & 7.14 &0.90 &5.49 & 0.84 \\

Super SloMo~\cite{jiang2017super}	&{2.51} 	 &0.59  &3.66 	 &0.72   	&\first{2.91} 		 &\second{0.74}  	&{5.05} 	  &	0.98 	&9.56  &	0.94 &5.37 	 &	{0.96}  	&6.69 & 0.60  	&6.73 & 	{0.69} 	&{5.31}   &	\second{0.78} 	 \\

CtxSyn~\cite{niklaus2018context}	&\first{2.24} 	 &\first{0.50} &\first{2.96} 	 &\first{0.55}   	&{4.32} 		 &{1.42}  	&\first{4.21} 	  &	\second{0.87}	&9.59  &	0.95 &5.22 	 &	{0.94}  	&7.02 & 0.68  	&6.66 & 	{0.67} 	&{5.28}   &	{0.82} 	 \\

\OurTPAMIs~\cite{bao2018MEMC-Net} & 2.47 & 0.60  & 3.49 &0.65 &4.63 &1.42 &4.94 &0.88& \second{8.91} &\second{0.93}& \first{4.70} & \second{0.86} & \second{6.46} &\second{0.66} & \second{6.35} &\first{0.64}& \second{5.24} & 0.83 \\

\Ours~(Ours) & \second{2.38} & 0.58  & \second{3.28} & \second{0.60} & \second{3.32} &\first{0.69} & \second{4.65} & \first{0.86}& \first{7.88} &\first{0.87}& \second{4.73} & \first{0.85} & \first{6.36} & \first{0.59} & \first{6.25} &\second{0.66}& \first{4.86} & \first{0.71} \\
\bottomrule
\end{tabular} 
\vspace{-10pt}
\end{table*}
	\begin{figure*}
	\footnotesize
	\centering
	\renewcommand{\tabcolsep}{1.0pt} 
	\renewcommand{\arraystretch}{1.0} 
	\begin{tabular}{ccccccccc}
%

 \multirow{2}{*}[3.05em]{	
\includegraphics[width=0.15\linewidth]{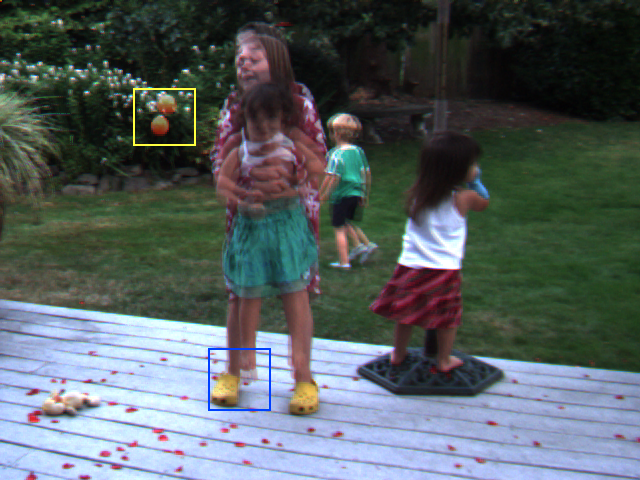}
}
&
\includegraphics[width=0.10\linewidth]{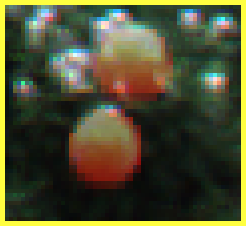} &
\includegraphics[width=0.10\linewidth]{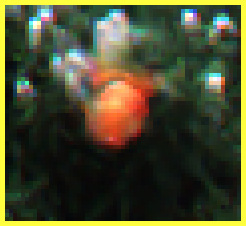}&
\includegraphics[width=0.10\linewidth]{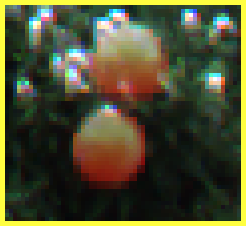} &
\includegraphics[width=0.10\linewidth]{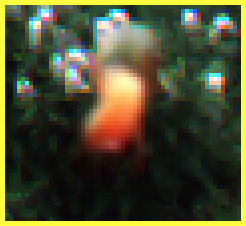}&
\includegraphics[width=0.10\linewidth]{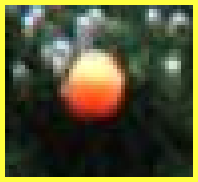}&
\includegraphics[width=0.10\linewidth]{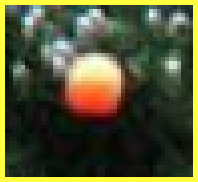}&
\includegraphics[width=0.10\linewidth]{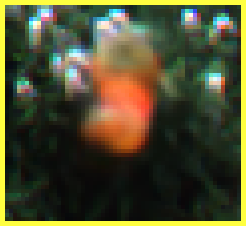} &
\includegraphics[width=0.10\linewidth]{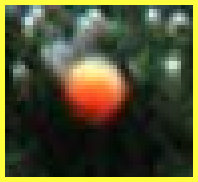}
\\	

&
\includegraphics[width=0.10\linewidth]{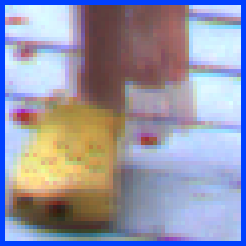} &
\includegraphics[width=0.10\linewidth]{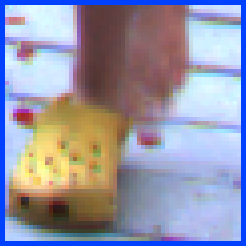}&
\includegraphics[width=0.10\linewidth]{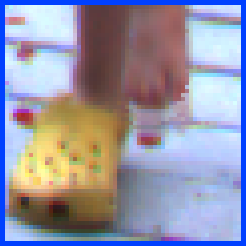} &
\includegraphics[width=0.10\linewidth]{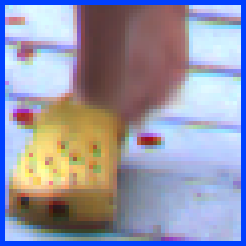}&
\includegraphics[width=0.10\linewidth]{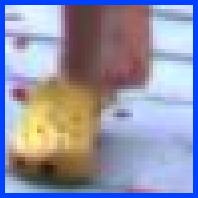}&
\includegraphics[width=0.10\linewidth]{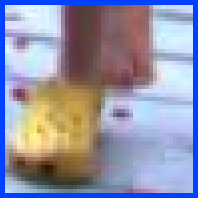}&
\includegraphics[width=0.10\linewidth]{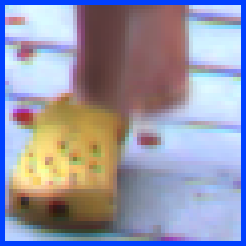} &
\includegraphics[width=0.10\linewidth]{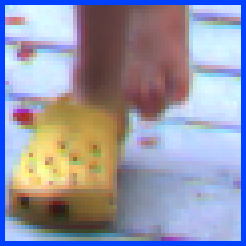} 
\\		
		&
			Inputs &
			ToFlow &
			EpicFlow&
			SepConv-$L_1$&
			Super SloMo &
			CtxSyn &
			\OurTPAMIs &
			\Ours~(Ours)\\

		\end{tabular}
	\vspace{-5pt}
	\caption{
	 \textbf{Visual comparisons on the Middlebury \textsc{Evaluation} set.}
	 The proposed method reconstructs a clear shape of the ball and restores more details on the slippers and foot.
	}
\vspace{-10pt}
\label{fig:Middlebury} 
\end{figure*}

	\begin{table*}[h!]
 \caption{
 	\textbf{Quantitative comparisons on the UCF101, Vimeo90K, HD, and Middlebury \textsc{Other} datasets.}
 	The numbers in \first{red} and \second{blue} indicate the best and second best performance.
 	We also compare the model parameters and runtime of each method.
 }
 \label{tab:UCF101_Vimeo90K_MB}
 \vspace{-5pt}
\footnotesize
\centering
\begin{tabular}{rccccccccc}
\toprule
\multirow{2}{*}[-0.28em]{Method}  &
\multirow{2}{*}{\thead{\#Parameters \\ (million)}} &
\multirow{2}{*}{\thead{Runtime \\ (seconds)}} 
&\multicolumn{2}{c}{UCF101~\cite{soomro2012ucf101}} &\multicolumn{2}{c}{Vimeo90K~\cite{xue2017video}} & Middlebury~\cite{baker2011database} &\multicolumn{2}{c}{HD~\cite{bao2018MEMC-Net}}   \\

\cmidrule(l{7pt}r{7pt}){4-5}
\cmidrule(l{7pt}r{7pt}){6-7}
\cmidrule(l{5pt}r{5pt}){8-8}
\cmidrule(l{7pt}r{7pt}){9-10}
&&	&PSNR & SSIM 	&PSNR & SSIM	& 
	IE &PSNR &SSIM\\
	\addlinespace[-1pt]
\midrule
SPyNet~\cite{ranjan2017optical}&1.20  & 0.11  &33.67 & 0.9633 &31.95 & 0.9601& 2.49 & ---& ---\\
EpicFlow~\cite{revaud2015epicflow}& --- &8.80 & 33.71 & 0.9635 &32.02&0.9622& 2.47  & ---& ---\\
MIND~\cite{long2016learning}& 7.60 &0.01 &  33.93 & 0.9661 & 33.50& 0.9429 & 3.35 &--- & ---\\ 
DVF~\cite{liu2017video}&1.60 & 0.47 &34.12 & 0.9631 & 31.54 &  0.9462 & 7.75  & ---& --- \\ 
ToFlow~\cite{xue2017video}&1.07 &0.43 & 34.58 & 0.9667  &33.73& 0.9682 &2.51 &29.37 & 0.8772 \\
SepConv-$L_f$~\cite{niklaus2017videoSepConv}& 21.6&0.20  & 34.69 & 0.9655  & 33.45& 0.9674& 2.44  & 30.61 & 0.8978 \\
SepConv-$L_1$~\cite{niklaus2017videoSepConv} & 21.6 & 0.20 & 34.78 & 0.9669  &33.79& {0.9702}& 2.27  &30.87 &0.9077 \\
\OurTPAMIs~\cite{bao2018MEMC-Net}  & 70.3 & 0.12 & \second{34.96} & \second{0.9682} &\second{34.29}& \second{0.9739} &\second{2.12}  &\second{31.39}&\second{0.9163} \\ 

\Ours~(Ours)  & 24.0 & 0.13 & \first{34.99} & \first{0.9683} &\first{34.71}& \first{0.9756} &\first{2.04}  &\first{31.64} & \first{0.9205}\\ 
\bottomrule
\end{tabular} 
\vspace{-5pt}
\end{table*}
	\begin{figure*}[h!]
	\footnotesize
	\centering
	\renewcommand{\tabcolsep}{0.8pt} 
	\renewcommand{\arraystretch}{0.5} 
	\begin{tabular}{cccccccccc}
\includegraphics[width=0.095\linewidth]{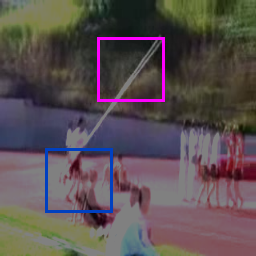}&
\includegraphics[width=0.095\linewidth]{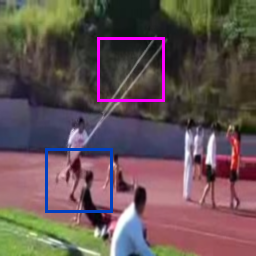}&
\includegraphics[width=0.095\linewidth]{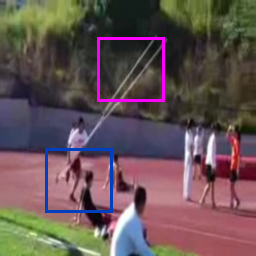}&
\includegraphics[width=0.095\linewidth]{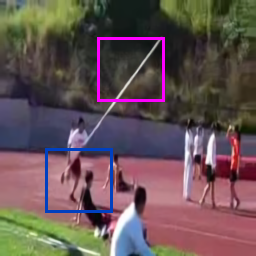}&
\includegraphics[width=0.095\linewidth]{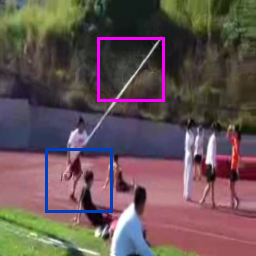}&
\includegraphics[width=0.095\linewidth]{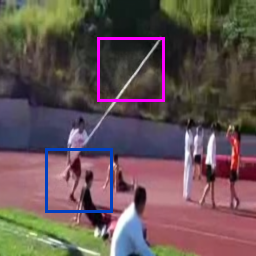}&
\includegraphics[width=0.095\linewidth]{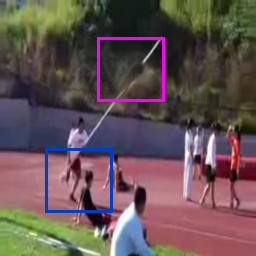}&
\includegraphics[width=0.095\linewidth]{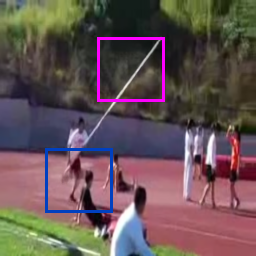}&
\includegraphics[width=0.095\linewidth]{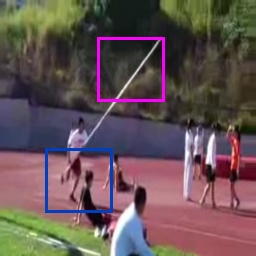}&
\includegraphics[width=0.095\linewidth]{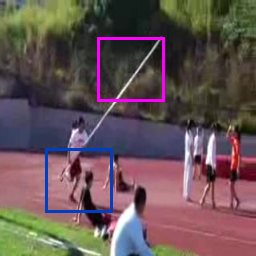}\\

\includegraphics[width=0.095\linewidth]{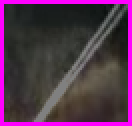}&
\includegraphics[width=0.095\linewidth]{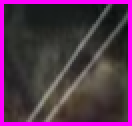}&
\includegraphics[width=0.095\linewidth]{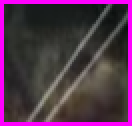}&
\includegraphics[width=0.095\linewidth]{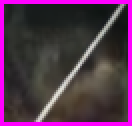}&
\includegraphics[width=0.095\linewidth]{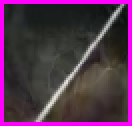}&
\includegraphics[width=0.095\linewidth]{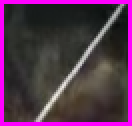}&
\includegraphics[width=0.095\linewidth]{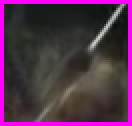}&
\includegraphics[width=0.095\linewidth]{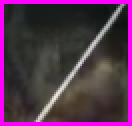}&
\includegraphics[width=0.095\linewidth]{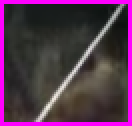}&
\includegraphics[width=0.095\linewidth]{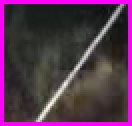}\\

\includegraphics[width=0.095\linewidth]{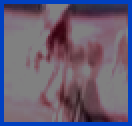}&
\includegraphics[width=0.095\linewidth]{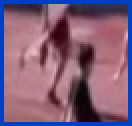}&
\includegraphics[width=0.095\linewidth]{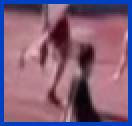}&
\includegraphics[width=0.095\linewidth]{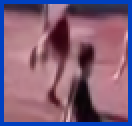}&
\includegraphics[width=0.095\linewidth]{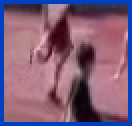}&
\includegraphics[width=0.095\linewidth]{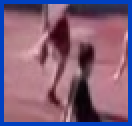}&
\includegraphics[width=0.095\linewidth]{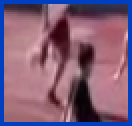}&
\includegraphics[width=0.095\linewidth]{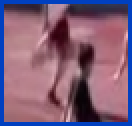}&
\includegraphics[width=0.095\linewidth]{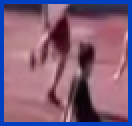}&
\includegraphics[width=0.095\linewidth]{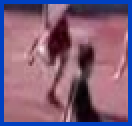}
\\				 	
Inputs&
SPyNet &
EpicFlow&
MIND& 
DVF&
ToFlow& 
SepConv-$L_1$&
\OurTPAMIs& 
\Ours~(Ours) &
Ground-truth\\				
\end{tabular}
	\vspace{-5pt}
	\caption{
		\textbf{Visual comparisons on the UCF101 dataset~\cite{soomro2012ucf101}.}
		The proposed method aligns the content (e.g., the pole) well and restores more details on the man's leg.
	}
	\label{fig:UCF101} 
	\vspace{-10pt}
\end{figure*}

	We show the quantitative results in~\tabref{context_ablation} and compare the interpolated images in~\figref{effect_of_context}.
	Without using any contextual information, the model does not perform well and generates blurred results.
	By introducing the contextual features, e.g., the pre-trained \textit{conv1} features or depth maps, the performance is greatly improved.
	We further demonstrate that the \emph{learned} contextual features, especially the learned hierarchical features, lead to a substantial improvement on the Vimeo90K and the Middlebury datasets.
	The model using both the depth maps and learned hierarchical features also generates sharper and clearer content.

\begin{table}
	\caption{
		\textbf{Comparisons with MEMC-Net~\cite{bao2018MEMC-Net} on parameter and runtime}.
		We list the parameters (million) and runtime (seconds) of each sub-module in the MEMC-Net and the proposed model.
		}
	\vspace{-5pt}
	\label{tab:detailtime}
	\footnotesize
	\centering
	\begin{tabular}{ccccc}
		\toprule 	
\multirow{2}{*}[-0.28em]{Sub-module}	&\multicolumn{2}{c}{\OurTPAMIs~\cite{bao2018MEMC-Net}} & \multicolumn{2}{c}{\Ours{} (Ours)}\\
			\cmidrule(l{2pt}r{2pt}){2-3}
			\cmidrule(l{2pt}r{2pt}){4-5}
			&\#Parameters &Runtime &\#Parameters &Runtime \\
			\addlinespace[-1pt]
			\midrule
Depth &---&--- &  5.35 & 0.043 \\
 
Flow   & 38.6 & 0.024 & 9.37 &0.074 \\
Context & 0.01 & 0.002 & 0.16 & 0.002 \\
Kernel  & 14.2 & 0.008 & 5.51 & 0.004 \\
Mask  &   14.2 & 0.008 & --- & --- \\
Synthesis  &  3.30 & 0.080& 3.63 & 0.002 \\
\midrule
Total	 & 70.3 & 0.122 & 24.0 & 0.125 \\
\bottomrule
	\end{tabular} 
	\vspace{-5pt}
\end{table}

\begin{figure}
	\footnotesize
	\centering
	\renewcommand{\tabcolsep}{1pt} 
	\begin{tabular}{ccc}

			\includegraphics[width=0.32\linewidth]{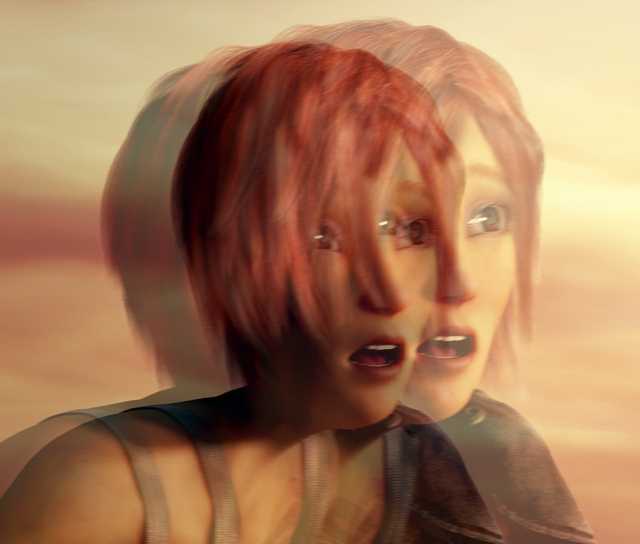}&
			\includegraphics[width=0.32\linewidth]{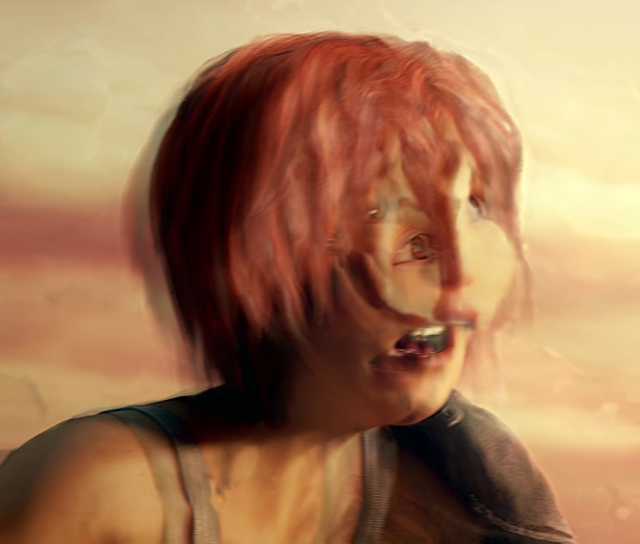}&
			\includegraphics[width=0.32\linewidth]{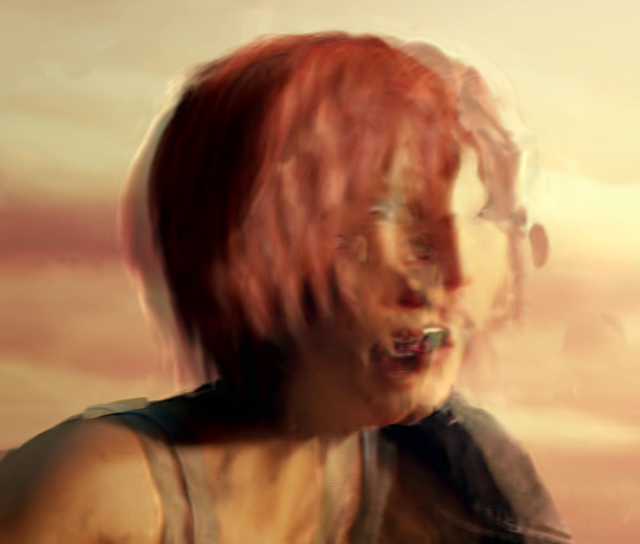}\\	
			Overlayed inputs&
			SepConv-$L_f$&
			SepConv-$L_1$
			\\
			\includegraphics[width=0.32\linewidth]{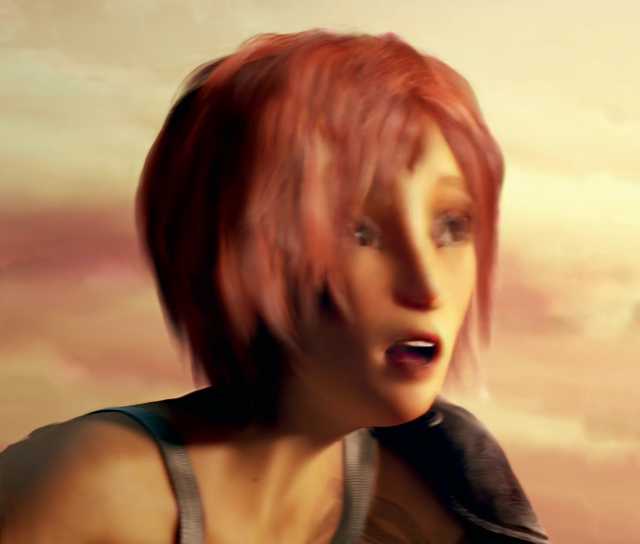} &
			\includegraphics[width=0.32\linewidth]{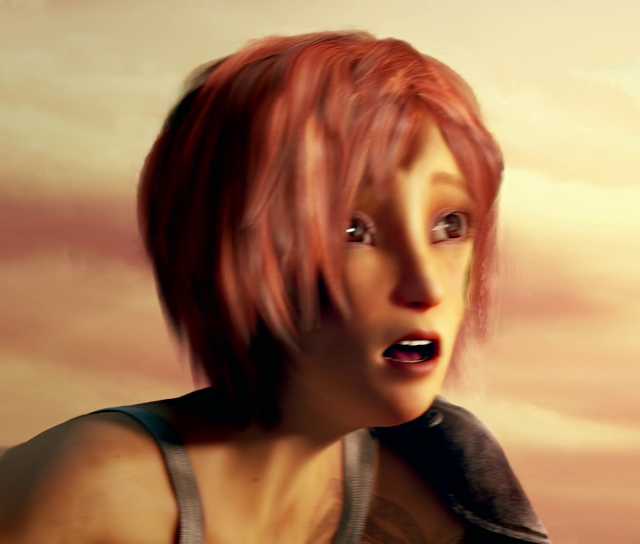} &
		    \includegraphics[width=0.32\linewidth]{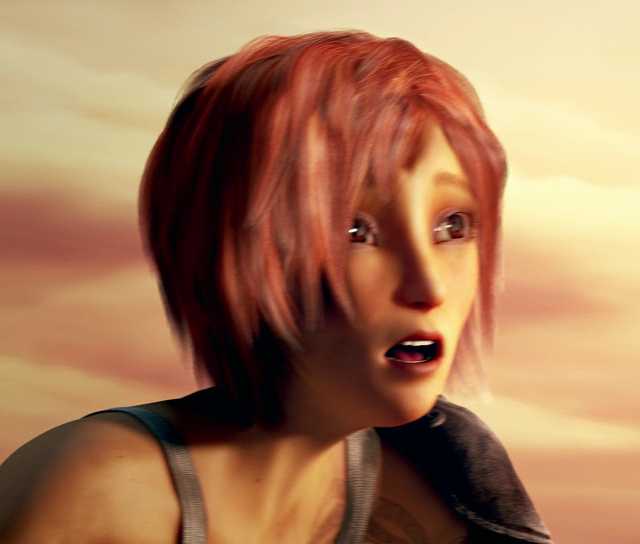}\\
		    
			\OurTPAMIs &
			\Ours~(Ours) &
			Ground-truth \\
		\end{tabular}
	\vspace{-5pt}
	\caption{
		\textbf{Visual comparisons on the HD dataset~\cite{bao2018MEMC-Net}.}
		The SepConv~\cite{niklaus2017videoSepConv} method cannot align the content as the motion is larger than the size of interpolation kernels, e.g., $51 \times 51$.
		The proposed DAIN reveals more details on the hair and eyes than the state-of-the-art MEMC-Net~\cite{bao2018MEMC-Net}.
	}
	\label{fig:HD-Alley2} 
	\vspace{-10pt}
\end{figure}

	\subsection{Comparisons with State-of-the-arts}
	We evaluate the proposed DAIN against the following CNN-based frame interpolation algorithms: MIND~\cite{long2016learning}, DVF~\cite{liu2017video}, SepConv~\cite{niklaus2017videoSepConv}, CtxSyn~\cite{niklaus2018context}, ToFlow~\cite{xue2017video}, Super SloMo~\cite{jiang2017super} and MEMC-Net~\cite{bao2018MEMC-Net}.
	In addition, we use the algorithm of Baker~\etal~\cite{baker2011database} to generate interpolation results for two optical flow estimation algorithms, EpicFlow~\cite{revaud2015epicflow} and SPyNet~\cite{ranjan2017optical}, for comparisons.

	In~\tabref{Middlebury}, we show the comparisons on the \textsc{Evaluation} set of the Middlebury benchmark~\cite{baker2011database}, which are also available on the Middlebury website.
	The proposed model performs favorably against all the compared methods.
	At the time of submission, our method ranks $1^\mathrm{st}$ in terms of NIE and $3^\mathrm{rd}$ in terms of IE among all published algorithms on the Middlebury website.
	We show a visual comparison in~\figref{Middlebury}, where the EpicFlow~\cite{revaud2015epicflow}, ToFlow~\cite{xue2017video}, SepConv~\cite{niklaus2017videoSepConv} and MEMC-Net~\cite{bao2018MEMC-Net} methods produce ghosting artifacts on the balls or foot.
	In contrast, the proposed method reconstructs a clear shape of the ball.
	Compared to the CtxSyn~\cite{niklaus2018context} and Super SloMo~\cite{jiang2017super} methods, our approach generates more details on the slippers and foot.

	In~\tabref{UCF101_Vimeo90K_MB}, we provide quantitative performances on the UCF101~\cite{soomro2012ucf101}, Vimeo90K~\cite{xue2017video}, HD~\cite{bao2018MEMC-Net}, and Middlebury~\cite{baker2011database} \textsc{Other} set.
	Our approach performs favorably against existing methods for all the datasets, especially on the Vimeo90K~\cite{xue2017video} dataset with a 0.42dB gain over MEMC-Net~\cite{bao2018MEMC-Net} in terms of PSNR.

	In~\figref{UCF101}, the SPyNet~\cite{ranjan2017optical}, EpicFlow~\cite{revaud2015epicflow} and SepConv~\cite{niklaus2017videoSepConv} methods cannot align the pole well and thus produce ghosting or broken results.
	The MIND~\cite{long2016learning}, DVF~\cite{liu2017video}, ToFlow~\cite{xue2017video} and MEMC-Net~\cite{bao2018MEMC-Net} methods generate blurred results on the man's leg.
	In contrast, the proposed method aligns the pole well and generates clearer results.
	In~\figref{HD-Alley2}, we show an example from the HD dataset.
	The SepConv~\cite{niklaus2017videoSepConv} method cannot align the content at all as the motion is larger than the size of the interpolation kernels (e.g., $51 \times 51$).
	Compared to the MEMC-Net~\cite{bao2018MEMC-Net}, our method restores clearer details on the hair and face (e.g., eyes and mouth).
	Overall, the proposed DAIN generates more visually pleasing results with fewer artifacts than existing frame interpolation methods.
	%
	In our supplementary materials, we demonstrate that our method can generate arbitrary intermediate frames to create $10\times$ slow-motion videos.
	More image and video results are available in our project website.

	We also list the number of model parameters and execution time (test on a $640\times 480$ image) of each method in~\tabref{UCF101_Vimeo90K_MB}.
	The proposed model uses a similar amount of parameters as the SepConv~\cite{niklaus2017videoSepConv} but runs faster.
	Compared to the MEMC-Net~\cite{bao2018MEMC-Net}, we use $69\%$ fewer parameters (see the detailed comparison of the sub-modules in~\tabref{detailtime}) and achieve better performance.

\begin{figure}
	\footnotesize
	\centering
	\renewcommand{\tabcolsep}{1pt} 
	\begin{tabular}{cccc}
		
		\includegraphics[width=0.285\linewidth]{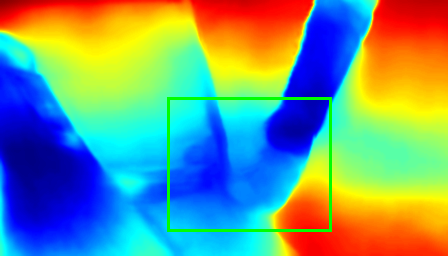}&
		\includegraphics[width=0.2\linewidth]{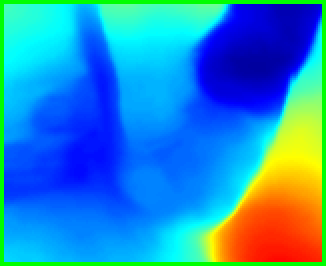}&
		\includegraphics[width=0.285\linewidth]{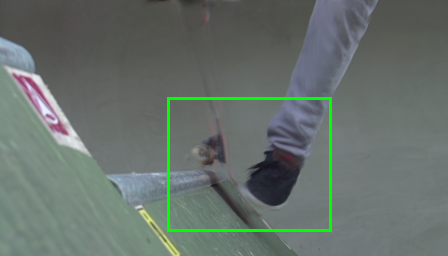} &
		\includegraphics[width=0.2\linewidth]{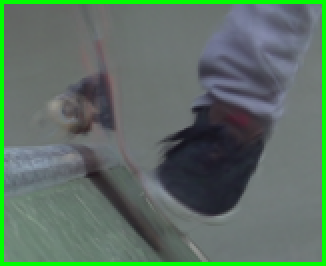}\\
		
	\multicolumn{2}{c}{Depth map} & \multicolumn{2}{c}{ToFlow}\\
	\addlinespace[2pt]
		
		\includegraphics[width=0.285\linewidth]{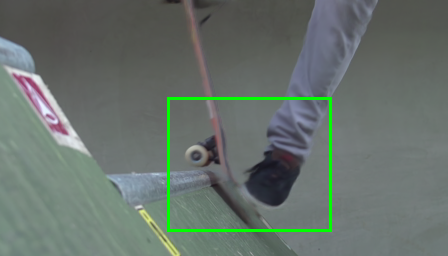} &
		\includegraphics[width=0.2\linewidth]{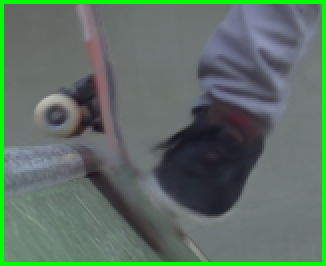} &
		\includegraphics[width=0.285\linewidth]{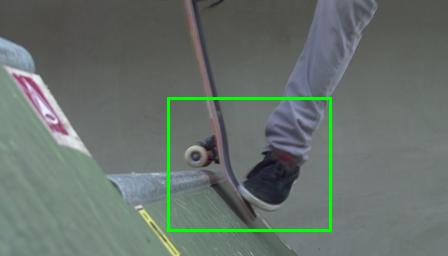}&
		\includegraphics[width=0.2\linewidth]{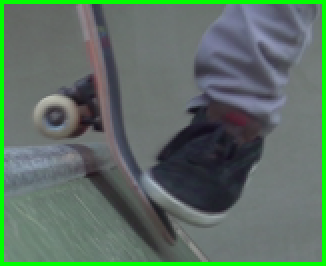}\\			
			\multicolumn{2}{c}{DAIN (Ours)}  & 	\multicolumn{2}{c}{Ground-truth} \\
	\end{tabular}
	\vspace{-5pt}
	\caption{
		\textbf{Limitations of the proposed method.}
		When the depth maps are not estimated well, our method tends to generate blurred results and less clear boundaries.
	}
	\label{fig:discussion} 
	\vspace{-10pt}
\end{figure}

	\subsection{Discussions and limitations}
	The proposed method relies on the depth maps to detect the occlusion for flow aggregation.
	However, in some challenging cases, the depth maps are not estimated well and lead to ambiguous object boundaries, as shown in the highlight region of~\figref{discussion}.
	Our method generates blurred results with unclear boundaries (e.g., between the shoe and skateboard).
	However, compared to the ToFlow~\cite{xue2017video}, our method still reconstructs the skateboard well.
	While our current model estimates depth from a single image, it would be beneficial to obtain more accurate depth maps by jointly estimating the depth from the two input frames or modeling the consistency between optical flow and depth~\cite{zou2018df}.

	\section{Conclusion}
	In this work, we propose a novel depth-aware video frame interpolation algorithm, which explicitly detects the occlusion using the depth information.
	We propose a depth-aware flow projection layer that encourages sampling of closer objects than farther ones.
	Furthermore, we exploit the learned hierarchical features and depth maps as the contextual information to synthesize the intermediate frame.
	The proposed model is compact and efficient.
	Extensive quantitative and qualitative evaluations demonstrate that the proposed method performs favorably against existing frame interpolation algorithms on diverse datasets.
	The state-of-the-art achievement from the proposed method sheds light for future research on exploiting the depth cue for video frame interpolation.

	%
	\small{
	\Paragraph{Acknowledgment}. 
	This work was supported in part by National Key Research and Development Program of China (2016YFB1001003), NSFC (61771306), Natural Science Foundation of Shanghai (18ZR1418100), Chinese National Key S\&T Special Program (2013ZX01033001-002-002), Shanghai Key Laboratory of Digital Media Processing and Transmissions (STCSM 18DZ2270700 and 18DZ1112300). It was also supported in part by NSF Career Grant (1149783) and gifts from Adobe, Verisk, and NEC.
}

	\newpage
	{\small
		\bibliographystyle{ieee}
		\bibliography{cvpr19_frame_interp_v6_arxiv}
	}
	
\end{document}